\documentclass[11pt]{article}
\usepackage[margin=1in]{geometry}
\usepackage{amsmath,amssymb,bm}
\usepackage{graphicx}
\usepackage{subcaption}
\usepackage[section]{placeins}
\usepackage{booktabs}
\usepackage{tabularx}
\usepackage{algorithm}
\usepackage{algpseudocode}
\usepackage{listings}
\usepackage{authblk}
\usepackage[round,authoryear]{natbib}
\usepackage{xcolor}
\definecolor{citeblue}{HTML}{1F5A94}
\usepackage[colorlinks=true,citecolor=citeblue,linkcolor=citeblue,urlcolor=citeblue,hyperfootnotes=false]{hyperref}
\usepackage{microtype}

\newcommand{\repositoryurl}{https://github.com/Alexiush/fleet}

\let\plaincitep\citep
\let\plaincitet\citet
\renewcommand{\citep}[1]{\textbf{\textcolor{citeblue}{\plaincitep{#1}}}}
\renewcommand{\citet}[1]{\textbf{\textcolor{citeblue}{\plaincitet{#1}}}}

\newcommand{\repositoryfootnote}{%
  \begingroup
  \renewcommand{\thefootnote}{}
  \footnotetext{\textbf{Algorithm source code and experiments:} \href{\repositoryurl}{\nolinkurl{\repositoryurl}}}
  \addtocounter{footnote}{-1}
  \endgroup
}

\numberwithin{figure}{section}
\numberwithin{table}{section}

\title{FLEET: From Logits Entropy to Enhanced Trajectories in Text Generation}
\author{Oleksii Streltsov\thanks{Work was done while Oleksii Streltsov was a master's student at Kharkiv National University of Radio Electronics.}\ }
\author{Oleksandra Vitko}
\affil{Department of Artificial Intelligence, Kharkiv National University of Radio Electronics, Kharkiv, Ukraine}
\date{}

\begin{document}
\maketitle
\thispagestyle{plain}
\repositoryfootnote

\begin{abstract}
Solutions based on large language models (LLMs) often rely on temperature sampling to improve accuracy and stability by aggregating multiple samples from the completion distribution. However, this memoryless approach is inherently suboptimal: because it lacks awareness of prior generations and their evaluations, it produces an increasing proportion of semantically duplicate answers as more samples are drawn, leading to diminishing returns. To address this limitation, we introduce \textbf{FLEET}, a novel method that integrates a memory mechanism into the generation process. FLEET represents each generation as a sparse trajectory through states whose entropy exceeds a predefined threshold and uses these trajectories to infer per-token utility scores that adjust the logits. Benchmark evaluations demonstrate that FLEET achieves the same accuracy as the repeated sampling baseline, with a 3x speedup, and substantially improves accuracy on complex coding tasks (LiveCodeBench Pass@32 increases from 59.9\% to 66.2\%) under the same budget. Furthermore, in the greedy-decoding configuration evaluated here, the approach is deterministic and uses a single calibration pass to derive its principal hyperparameters, requiring only minimal modifications to existing LLM pipelines. 

\end{abstract}
\section{Introduction}

Large language models (LLMs) inherently operate under conditions of high uncertainty. Their capacity to function effectively in such environments stems from their robust predictive performance, which is largely maintained during autoregressive decoding \citep{he2024law}. However, the sequential nature of LLM generation implies that a localized failure at a pivotal step can induce cascading errors, ultimately derailing the entire reasoning trajectory. Consequently, there remains a pronounced disparity between the models' high proficiency in isolated next-token prediction and their success rates in multi-step, autonomous problem-solving (agentic) tasks \citep{laban2025llms}.

To address this limitation, recent research has increasingly focused on test-time scaling, a paradigm that leverages additional computational resources during inference to mitigate autoregressive failures \citep{zhang2025survey}. Such methods include aggregating multiple model completions to identify the optimal or most frequent response \citep{zhou2025theoretical}, building iterative self-refinement frameworks \citep{shinn2023reflexion} or training the models to intrinsically evaluate their intermediate outputs, enabling them to dynamically allocate supplementary inference compute based on task complexity \citep{deepseekai2025deepseekr1}.

In this work, we specifically focus on the paradigm of sampling multiple completions. First, prior literature demonstrates that sampling remains one of the most cost-effective and accessible test-time interventions \citep{snell2024scaling}. Consequently, methodological advancements in this domain are highly generalizable, yielding immediate performance benefits across virtually any decoder-only generative architecture \citep{yenduri2023generative}. Second, in contrast to alternative methods that strictly depend on the intrinsic capabilities of the model, sampling affords fine-grained, inherently task-agnostic control over the generation trajectory. Finally, sampling mechanisms constitute a foundational component of the post-training alignment pipelines utilized in the development of modern large language models \citep{ouyang2022training}.

The conventional baseline for token selection in large language models is greedy decoding, which deterministically selects the candidate token associated with the highest conditional probability at each autoregressive step. However, this deterministic paradigm is inherently ill-suited for test-time scaling strategies, as it yields zero sample variance and produces identical completion trajectories across repeated invocations. To induce diversity into the generation process, stochastic temperature sampling is commonly employed. Its temperature-controlled distribution is related to the Boltzmann sampling used in early stochastic neural models \citep{ackley1985learning}. Rather than selecting the distribution mode, temperature sampling scales the unnormalized model logits by a temperature parameter $T > 0$ prior to applying the softmax operator, thereby constructing a re-scaled probability distribution from which subsequent tokens are stochastically drawn. This mechanism enables the generation of distinct completion trajectories while maintaining probability mass aligned with the model's underlying likelihood estimates \citep{brown2024large}:

\begin{equation}
x_{\mathrm{next}} = \arg\max_{x \in \mathcal{V}} P(x)
\end{equation}
\begin{equation}
P(x_i) = \frac{\exp(l_i)}{\sum_{x_j \in \mathcal{V}} \exp(l_j)}
\end{equation}
\begin{equation}
P_T(x_i) = \frac{\exp(l_i / T)}{\sum_{x_j \in \mathcal{V}} \exp(l_j / T)}
\end{equation}

We identify key structural limitations in conventional sampling approaches:

\paragraph{a) Sample Inefficiency}

Consider a generation containing a small subset of tokens that are crucial to task success, which we call branching points. Success requires selecting an appropriate branch at each such point. When the optimal token does not correspond to the highest-likelihood mode, standard sampling strategies continuously over-allocate probability mass to unviable candidates. As the frequency of these branching points increases, the joint probability of sampling a globally correct trajectory decays exponentially. Simply increasing the temperature parameter fails to resolve this issue. While a higher temperature elevates the likelihood of selecting non-modal optimal tokens, it uniformly inflates variance across all generation steps. This indiscriminate entropy injection destabilizes generation at otherwise stable steps, frequently degrading overall success rates. Consequently, temperature scaling does not constitute a principled solution to sample inefficiency; rather, it acts merely as a static control parameter to navigate the trade-off between exploration and precision.

\paragraph{b) Lack of Selective Exploration}

In any given state, only a narrow subset of the vocabulary represents valid or task-relevant continuations. Furthermore, a substantial fraction of autoregressive steps are predominantly syntactic or structural (e.g., punctuation, functional words, or deterministic code syntax), operating under low entropy. Ideally, stochastic sampling should be restricted to high-entropy decision points and semantically meaningful actions, rather than applied uniformly across all generation steps. To partially mitigate the inclusion of unviable tokens, heuristic logit truncation strategies are commonly integrated into the sampling pipeline:

\begin{itemize}
\item Top-$k$ sampling, which restricts the candidate vocabulary to a fixed subset of $k$ tokens with the highest conditional probabilities \citep{fan2018hierarchical};
\item Top-$p$ (nucleus) sampling, which dynamically selects the minimal set of candidate tokens whose cumulative probability mass exceeds a threshold parameter $p$ \citep{holtzman2019curious};
\item Min-$p$ sampling, which truncates candidate tokens whose conditional probability falls below a dynamic threshold scaled relative to the likelihood of the distribution mode $P_{\max}$ \citep{nguyen2024turning}:
\end{itemize}

\begin{equation}
\mathcal{V}'_{top-k} = \{ x \in \mathcal{V} \mid \text{rank}(P(x)) \leq k \}
\end{equation}
\begin{equation}
\mathcal{V}'_{top-p} = \arg\min_{\mathcal{S} \subset \mathcal{V}} \left( |\mathcal{S}| \right) \quad \text{subject to} \quad \sum_{x \in \mathcal{S}} P(x) \geq p
\end{equation}
\begin{equation}
\mathcal{V}'_{min-p} = \{ x \in \mathcal{V} \mid P(x) \geq p \cdot P_{\max} \}
\end{equation}

\paragraph{c) Low Robustness and Poor Cross-Task Generalization}

The concurrent deployment of temperature scaling and the aforementioned logit-truncation strategies yields a highly parameterized search space (encompassing $T$, $k$, $p$, and $p_{\text{min}}$). Because these parameters lack a unified, principled theoretical foundation, their selection relies almost entirely on ad hoc empirical tuning. Consequently, optimal hyperparameter configurations exhibit significant sensitivity to prompt design, model architecture, and task complexity, failing to generalize across diverse downstream domains. As a result, practitioners rarely work with optimally tuned models and instead rely on domain-oriented configurations.

We are interested in an alternative strategy that is free from these limitations. We investigate existing logit-processing strategies beyond greedy decoding and temperature sampling and analyze how they mitigate the limitations described above. We find that current solutions can mitigate limitations (b) and (c), but not (a). We attribute this shortcoming to the fact that these sampling methods are unaware of the rewards associated with generated completions and therefore cannot distinguish between states in which they should explore and those in which they should exploit. As a solution to this problem, we propose a lightweight memory mechanism, an algorithm that uses it to sample better completions and evaluate this setup on two benchmarks with verifiable tasks.

In summary, our key contributions are as follows:
\begin{itemize}
\item We introduce Vector Disjoint Set Union (VectorDSU), an online data structure that maps hidden states into unified search states, preventing trajectory duplication and preserving state utility history.
\item We propose FLEET, a memory-augmented, deterministic search paradigm that replaces memoryless temperature sampling with targeted exploration of the completion space.
\item We provide an empirical analysis of FLEET performance comparing it to repeated sampling on mathematical and coding problems using both ground truth verifiers and reward models. We confirm that on these tasks our approach scales better as its accuracy is always higher under the same budget.
\end{itemize}

\section{Related Work}

One widely used alternative to greedy decoding is beam search, particularly in neural machine translation \citep{wu2016googles}. At each decoding step, beam search expands the retained partial sequences with candidate next tokens and keeps a fixed number of the highest-scoring hypotheses. The final output is typically selected according to accumulated sequence log-probability, often with a length adjustment. This deterministic search procedure can improve sequence-level consistency without requiring an external completion evaluator. However, it does not resolve the limitations described above because its search remains guided by the model's likelihood estimates; when high-likelihood continuations are incorrect, beam search may systematically favor them.

To address the inherent limitations of static decoding, several adaptive sampling algorithms have been introduced. These methods generally share the following strategy \citep{zhu2023hot} \citep{chang2025real}:

\begin{enumerate}
\item State Salience Detection: An auxiliary scoring mechanism or heuristic evaluates the generation context at step $i$ to identify critical or high-entropy decoding states (e.g., decision nodes characterized by high prediction uncertainty or task relevance).
\item Dynamic Parameter Modulation: Decoding hyperparameters (such as temperature $T$ or truncation thresholds $p$, $k$) are adaptively re-scaled as a function of the detected state properties, thereby balancing exploration and exploitation on a step-by-step basis.
\end{enumerate}

Formally, a representative adaptive policy that dynamically elevates the sampling temperature $T_i$ when encountering a high-salience state $s_i$ can be expressed as:

\begin{equation}
T_i = \begin{cases} a = f(s_i) & \text{if } s_i \in \mathcal{S}', \\ b & \text{otherwise} \end{cases} \text{ where } {a > b}
\end{equation}

While state-dependent modulation provides a principled mechanism to mitigate the lack of selective exploration (limitation b), it does not fully resolve the structural intricacies of test-time scaling. To address hyperparameter brittleness and poor generalization (limitation c), recent approaches have integrated learnable modules capable of governing dynamic parameter adjustments \citep{dang2026temperature}. Although these methods still rely on data-driven optimization, the tuning process is coupled directly with the objective of the target task, effectively bypassing the necessity for ad-hoc, task-agnostic manual searches.

Nevertheless, the fundamental vulnerability -- sample inefficiency at critical decision nodes (limitation a) -- persists. The persistence of this issue indicates that dynamically searching for an ``optimal'' temperature is a fundamentally misaligned objective; scalar logit adjustments cannot selectively amplify specific valid tokens without concurrently inflating the variance of the entire distribution. Existing adaptive-temperature results are task-dependent, and their generality across broader domains remains unclear.

\section{FLEET Algorithm}

In the context of test-time scaling, the fundamental objective is rarely to faithfully approximate the model's predictive distribution; rather, it is to isolate optimal, high-reward trajectories from within a vast hypothesis space. Standard temperature sampling, however, is inherently memoryless. It fails to leverage the evaluative feedback, derived from either verifiable task environments or learned preference models, that is usually used to evaluate the scaling process. Consequently, it has no concept of exploitation. Integrating a historical memory mechanism transforms this paradigm, enabling the decoding process to be steered via structured, search-like dynamics rather than blind stochasticity. By retaining evaluative information across generation iterations, such an approach can systematically navigate the combinatorial explosion of the token space, effectively bypassing the inefficiencies associated with temperature scaling. Consequently, we argue that resolving the aforementioned decoding limitations necessitates an algorithm designed to search the model's completion space systematically, rather than merely sample from it.

To resolve the limitations outlined above, we introduce FLEET (From Logits Entropy to Enhanced Trajectories) -- a memory-augmented sampling framework designed to systematically mitigate the structural inefficiencies of standard and adaptive decoding. FLEET employs an information-theoretic heuristic based on distribution entropy to dynamically detect high-uncertainty decoding states, identifying them as pivotal branching points. These states are subsequently indexed within a graph-like data structure, mapping local state representations to metadata that tracks historical completion trajectories traversing through them. Rather than relying on global scalar hyperparameter adjustments, FLEET utilizes this memory to evaluate candidate paths in a manner analogous to Monte Carlo Tree Search (MCTS) \citep{browne2012survey}. By leveraging historical evaluative outcomes, the algorithm selectively applies targeted logit penalties to actions associated with suboptimal trajectories, redirecting probability mass toward more promising search branches. Furthermore, hyperparameter selection within FLEET is principled and data-driven, as opposed to the black-box optimization typically required by standard sampling pipelines.

Formally, the entropy heuristic is derived from the model's unnormalized logit output, from which we compute two complementary information-theoretic metrics: conditional entropy $H$ and varentropy $V$.

\begin{equation}
H(X) = - \sum_{x} p(x) \log p(x)
\end{equation}
\begin{equation}
V(X) = \sum_{x} p(x) (\log p(x) + H(X))^2
\end{equation}

Employing a single uncertainty metric is insufficient, as conditional entropy and varentropy describe complementary aspects of the probability distribution's shape:

\begin{itemize}
\item Conditional Entropy ($H$): Measures total distributional uncertainty. However, standard Shannon entropy is susceptible to false positives in the presence of semantically redundant tokens (e.g., synonym clusters or stylistic variations) and during structured reasoning steps (e.g., deterministic mathematical operations), where high entropy does not necessarily reflect true semantic divergence.
\item Varentropy ($V$): Quantifies the variance (dispersion) of log-probabilities around the mean entropy. Varentropy remains relatively invariant under broad, semantically uniform synonym distributions, but exhibits pronounced spikes during multimodal decision steps where probability mass is split across distinct, non-equivalent candidate clusters.
\end{itemize}

For numerical scaling within the calibration pipeline, both metrics are normalized relative to the model's embedding dimension.

Furthermore, empirical observations demonstrate that final-layer logits often fail to provide the most sensitive uncertainty signals due to late-stage probability smoothing. To capture sharper decision-making dynamics, we extract hidden representations $h^{(l)}$ from an intermediate layer $l < L$ and project them directly into the vocabulary space using the language model head $W_U$ -- an interpretability technique known as the Logit Lens \citep{nostalgebraist2020interpreting}:

\begin{equation}
\mathbf{l}_t^{(l)} = \mathbf{W}_U \mathbf{h}_t^{(l)} + \mathbf{b}_U
\end{equation}
\begin{equation}
P^{(l)}(x_i) = \frac{\exp(l_{t,i}^{(l)})}{\sum_{x_j \in \mathcal{V}} \exp(l_{t,j}^{(l)})}
\end{equation}

Because intermediate hidden states lie in a continuous vector space, identifying semantically equivalent decision nodes and aggregating historical trajectory statistics across generations constitutes a nontrivial clustering task. To dynamically map continuous representations to discrete equivalence classes, we introduce a special data structure termed Vector Disjoint Set Union (VectorDSU):

VectorDSU is motivated by an empirical property of representation alignment: above a calibrated threshold $\tau$, high cosine similarity $S_C$ between two hidden vectors is associated with bounded Kullback--Leibler (KL) divergence $D_{KL}$ between their projected probability distributions $P$ and $Q$ over the vocabulary $\mathcal{V}$:

\begin{equation}
S_C(\mathbf{h}_1, \mathbf{h}_2) = \frac{\mathbf{h}_1 \cdot \mathbf{h}_2}{\|\mathbf{h}_1\| \|\mathbf{h}_2\|}
\label{eq:cosine-similarity}
\end{equation}
\begin{equation}
D_{\mathrm{KL}}(P \parallel Q) = \sum_{x \in \mathcal{V}} P(x) \log \left( \frac{P(x)}{Q(x)} \right)
\end{equation}

This alignment enables the robust mapping of continuous hidden states into discrete topological clusters $\mathcal{C}_k$, serving as anchor points to which trajectory metadata is attached:

\begin{equation}
S_C(\mathbf{h}_1, \mathbf{h}_2) \geq \tau \implies D_{\mathrm{KL}}(P_1 \parallel P_2) \leq \epsilon, \quad \text{where } P_i = \operatorname{softmax}(W\mathbf{h}_i + \mathbf{b})
\label{eq:cosine-kl-operating-criterion}
\end{equation}

This operating criterion is selected empirically for the model and layer being calibrated. It is theoretically motivated by the use of normalized hidden states. The projection from intermediate hidden states to output probabilities consists of a linear transformation (the language model head) followed by a softmax nonlinearity. Because the softmax function is Lipschitz-continuous and invariant to uniform scalar shifts, angular proximity in the latent continuous space intrinsically constrains the statistical divergence of the resulting output distributions. The exact tightness of this bound depends on the spectral norm of the LM head weights, which in practice are tightly bound by regularization and initialization schemes.

\subsection{Online Hidden State to Search State Mapping via Vector Disjoint Set Union}

The proposed online clustering mechanism is inspired by the disjoint-set data structure, commonly referred to as Disjoint Set Union (DSU) \citep{galil1991data}. VectorDSU borrows DSU's representative-centric organization, but it does not retain an exact union--find graph over all high-dimensional vectors. Instead, it maintains compact component representatives and the trajectory metadata associated with the cluster of vectors that resolve to it. Its operation follows three structural principles:

\begin{itemize}
\item Canonical Representation: Each component maintains a representative vector. Like in DSU vector that resolves to that representative is considered a member. Thus, once a state is assigned, its metadata resolves to that component even though the high-dimensional member vector need not be retained.
\item Virtual Connectivity: Component membership is induced by the history of online assignments and merges, analogously to connectivity in DSU. It does not require every historical member to remain directly similar to the current representative.
\item Low-Memory Dynamic Union: VectorDSU merges component identifiers, representatives, and metadata without storing every member vector. The representative may remain fixed or be updated as a running centroid, depending on the configured variant.
\end{itemize}

For a new hidden state $\mathbf h$, VectorDSU uses cosine similarity to choose an existing representative $\mu_i$ or to create a new component. This is an online assignment rule rather than a requirement that every historical member remain directly related to the representative. Retaining only representative vectors and compact component metadata substantially reduces memory use; representative lookup can additionally be accelerated through parallelization or spatial partitioning.

\begin{equation}
C(\mathbf h)=\arg\max_i S_C(\mathbf h,\mu_i)
\quad\text{if}\quad
\max_i S_C(\mathbf h,\mu_i)\geq\tau;
\quad\text{otherwise create a new component.}
\end{equation}

Once a state is mapped to a cluster, the cluster functions as a memory node, recording historical trajectory metadata. Specifically, each cluster aggregates transition tuples reflecting the generation dynamics. For a given autoregressive step originating in state $\mathcal{C}_t$, the stored metadata comprises:

\begin{itemize}
\item the action (token) executed $a_t$;
\item the state $C_{t+1}$ reached upon executing action $a_t$;
\item the reward $R(s_t, a_t)$ accumulated along the trajectory path.
\end{itemize}

\begin{equation}
\mathcal{M}_{t} = \langle a_t, C_{t+1}, \bar{R}(s_t, a_t) \rangle
\end{equation}

Because state cluster retrieval relies on intermediate vector representations, FLEET assumes access to the model's internal activations and output probability distributions. This transparency enables the integration of Monte Carlo Tree Search (MCTS) to navigate the sequence space, specifically utilizing the predictor-guided Upper Confidence Bound applied to Trees (pUCT) formulation \citep{silver2017mastering}.

Standard UCT evaluates actions by balancing an empirical exploitation term $Q(s, a)$ -- defined as the average observed reward for executing action $a$ in state $s$ -- with an exploration bonus driven by relative visit counts and scaled by an exploration constant $c$. However, standard UCT assumes exhaustive local exploration, rendering it unsuited for vast vocabulary spaces. In contrast, pUCT incorporates an explicit prior policy $P(s, a)$ -- naturally supplied by the language model's unpenalized output probability distribution -- to bias search toward semantically viable candidates. Additionally, it rescales the exploration dynamics to support online, non-exhaustive tree expansion:

\begin{equation}
UCT(s, a) = Q(s, a) + c \sqrt{\frac{\ln N(s)}{N(s, a)}}
\end{equation}

\begin{equation}
pUCT(s, a) = Q(s, a) + c_{puct} P(s, a) \frac{\sqrt{N(s)}}{1 + N(s, a)}
\end{equation}

Here, $N(s,a)$ is the number of times action $a$ has been evaluated from state $s$, and $N(s)=\sum_a N(s,a)$ is the total visit count of that state.

Direct application of the standard pUCT formulation requires crucial modifications to accommodate the stochasticity and high dimensionality inherent to autoregressive language generation. Because a single token action $a$ evaluated from state $s$ may transition into a distribution of potential subsequent clusters, we evaluate action utility $Q(s, a)$ as the expectation of rewards across all known outcomes.

Furthermore, to mitigate the distortive effects of the model's predictive priors, we compute $P(s,a)$ by applying the resampling temperature $T_{\mathrm{resample}}$ to the logits before softmax, following Equation~(3), and then restrict the decision set to the $k$ most probable tokens. Let $\mathcal U(s)$ be the tokens in this top-$k$ set for which $N(s,a)=0$. These candidates are aggregated into an ``exploration action'' whose prior mass is $\sum_{a\in\mathcal U(s)}P(s,a)$. When this action is selected, the configured decoding rule chooses a concrete token from $\mathcal U(s)$. Thus, unexplored tokens remain available as a joint alternative, whereas known suboptimal tokens can be penalized individually.

Finally, rather than explicitly dictating the final token selection or replacing standard decoding pipelines, FLEET intervenes as a soft constraining mechanism. We preserve the configured decoding strategy, but apply a logit penalty $\lambda$ to empirically suboptimal tokens -- specifically, those that have been decoded and evaluated, yet fail to maximize the updated pUCT objective:

\begin{equation}
\hat{l}_{t,a} = l_{t,a} - \lambda \cdot \mathbb{I} \left[ N(s_t, a) > 0 \land a \neq \arg\max_{a'} pUCT(s_t, a') \right]
\end{equation}

Furthermore, the decoupled VectorDSU memory permits state-action statistics or priors derived from one worker or task to be supplied to other FLEET workers. This provides an interface for cross-session or cross-domain transfer, although the benefit of such transfer is not evaluated in the present experiments. In addition to the language model's localized predictive prior, the search policy can therefore incorporate a global domain prior when one is available.

In the absence of an external task-specific signal, explored actions receive a default multiplicative prior of 0.5, whereas the aggregated exploration action does not have one which can be interpreted as a prior of 1. This deliberately downscales evaluated actions relative to the unexplored candidate pool during early search; the value 0.5 is therefore an exploration bias rather than a learned domain probability.

\begin{equation}
\hat{l}_{t,a} = l_{t,a} - \lambda \cdot \mathbb{I} \left[ N(s_t, a) > 0 \land a \neq \arg\max_{a'} \left\{pUCT(s_t, a') \cdot prior(s_t, a')\right\} \right]
\end{equation}

In our experiments, as an experiment-specific heuristic, the logit penalty $\lambda$ was dynamically set to the maximum logit while we evaluated greedy decoding from the FLEET-processed scores.

\subsection{FLEET Search}

Synthesizing the components detailed above, the complete operational workflow of the proposed method is formalized in Algorithm~1.

\begin{algorithm}[!t]
\caption{FLEET: From Logits Entropy to Enhanced Trajectories}
\begin{algorithmic}[1]
\Require Generator $G$, evaluator $RM$, prompt $P$, budget $B$
\Require Decoding rule $\textsc{Decode}$ (greedy in our experiments)
\Require Layer $L$, entropy thresholds $(\tau_H,\tau_V)$, DSU threshold $\tau_{\mathrm{dsu}}$
\State Initialize VectorDSU $\mathcal{D}$ with threshold $\tau_{\mathrm{dsu}}$
\State Initialize FLEET worker $W$ linked to $\mathcal{D}$
\State $i \leftarrow 1$
\While{$W.\text{MaxReward()} < 1$ and $B > 0$}
  \State $Y \leftarrow \textsc{StartGeneration}(G,P)$ \Comment{Reset the trajectory}
  \State $W.\textsc{ResetTrajectory}()$
  \State $t \leftarrow 0$
  \While{$G$ is generating and $\textsc{LastToken}(Y) \neq \texttt{<EOS>}$}
    \State Obtain hidden state $\mathbf{h}_t^{(L)}$ from layer $L$
    \State Compute $H(P_t)$ and $V(P_t)$ from $\mathbf{h}_t^{(L)}$ via LogitLens
    \State $W.\textsc{UpdateUncertaintyBuffer}(H(P_t),V(P_t))$
    \If{$H(P_t) > \tau_H \land V(P_t) > \tau_V$}
      \State $C_t \leftarrow \mathcal{D}.\textsc{FindOrCreate}(\mathbf{h}_t^{(L)})$ \Comment{Map to a discrete state}
      \State $\mathbf{l}_t \leftarrow G.\textsc{GetLogits}()$
      \State $\hat{\mathbf{l}}_t \leftarrow \mathbf{l}_t - W.\textsc{CalculatePenalty}(C_t)$
      \State $a_t \leftarrow \textsc{Decode}(\hat{\mathbf{l}}_t)$ \Comment{Greedy in our experiments}
      \State $W.\textsc{RegisterAction}(C_t,a_t)$
    \Else
      \State $a_t \leftarrow \textsc{Decode}(G.\textsc{GetLogits}())$
    \EndIf
    \State $Y \leftarrow \textsc{Append}(Y,a_t)$
    \State $t \leftarrow t + 1$
  \EndWhile
  \State $R \leftarrow RM.\textsc{Evaluate}(Y)$
  \State $W.\textsc{BackpropagateReward}(R)$ \Comment{Update $\bar{R}$ and $N$ in $\mathcal{D}$}
  \State $\rho_i \leftarrow \min(i^{1/3},100)\%$
  \If{$W.\textsc{ThresholdHitRate}() < \rho_i$}
    \State $(\tau_H,\tau_V) \leftarrow W.\textsc{AdjustThresholds}(\rho_i)$
  \EndIf
  \State $B \leftarrow B - 1$
  \State $i \leftarrow i + 1$
\EndWhile
\end{algorithmic}
\end{algorithm}
To sustain search-like exploration dynamics, FLEET maintains a fixed-capacity running buffer of recent entropy--varentropy pairs and measures the percentage that satisfies the joint trigger $H>\tau_H \land V>\tau_V$. When this observed threshold-hit rate falls below the target $\rho(i)$, the entropy and varentropy thresholds are relaxed using the empirical distribution in the buffer so that the hit rate approaches the target. The target percentage follows a cube-root schedule:

\begin{equation}
\rho(i) = \min\!\left(i^{1/3}, 100\right)\%
\end{equation}

where $i$ denotes the search iteration index. This monotonic schedule gradually increases the target fraction of token states processed by FLEET. The buffer-based update permits the search to continue expanding even when the initial thresholds become too selective for later iterations.

The FLEET architecture introduces several other operational advantages:

\begin{itemize}
\item FLEET admits coordinated parallelization: workers can reserve or be assigned distinct high-scoring branches in the shared search memory, reducing overlap between concurrently generated trajectories. Quantifying the resulting parallel speedup remains future work.
\item Structural updates to VectorDSU and cluster-level penalty tables can be batched after each completed generation. During the next generation, only hidden-state lookup and application of the precomputed cluster penalties must occur online. This decoupled design ensures that the algorithm is fully compilable with highly optimized production inference backends that compile the computation graph.
\item Because the search operates as a soft constraint mechanism -- guiding generation through dynamic logit penalization rather than replacing the configured decoder -- it does not require changes to downstream evaluation beyond receiving a different generated sequence.
\item VectorDSU constructs token-level attributed trajectory data that records which probable actions were attempted, their empirical rewards, and how they compare with alternatives. This information is richer than a single sequence-level correctness label and may be useful for downstream Supervised Fine-Tuning (SFT) or Reinforcement Learning (RL), although its training value is not evaluated here.
\end{itemize}

We also acknowledge several limitations in the current implementation:

\begin{itemize}
\item Unlike standard sampling techniques that operate entirely post-hoc on the generated logits, our approach currently requires direct patching of the underlying model architecture. While it is theoretically possible to apply this method exclusively to the output logits, we expect that it will require significantly more memory and operations, possibly with degraded performance.
\item Our method can struggle when the model enters certain specific cognitive states, most notably those associated with induction or deep reflection. In inductive scenarios, defaulting to deterministic selection often mitigates the issue. However, for reflective states -- which are increasingly central to the performance of ``thinking'' or reasoning-focused models -- deterministic selection is actively detrimental. Consequently, our current method may not synergize well with advanced reasoning models that rely heavily on these internal reflective processes.
\item Finally, we observe that the generated text can occasionally exhibit high levels of noise, where structurally less important tokens are degraded into unintelligible sequences (gibberish). While this phenomenon of sampling noise is not unique to our specific method, it remains a practical limitation. A secondary, lightweight language model may mitigate primarily syntactic or formatting corruption by rewriting the generated output, but this mitigation is not evaluated in the present experiments.
\end{itemize}

\subsection{Hyperparameter Initialization}

FLEET requires the following hyperparameters to be set:

\begin{itemize}
\item $L$: The intermediate layer index from which continuous hidden states are extracted.
\item $\tau_H, \tau_V$: The baseline trigger thresholds for conditional entropy and varentropy, respectively.
\item $T_{\mathrm{resample}}, k$: The resampling temperature and the top-$k$ truncation bound utilized during prior policy evaluation.
\item $\tau_{\mathrm{dsu}}$: The critical cosine similarity threshold governing equivalence class formation within the VectorDSU.
\end{itemize}

Although FLEET lacks universal ``default'' settings -- a departure from conventional stochastic sampling methods -- its initialization process circumvents the need for exhaustive, iterative grid searches over validation datasets. Because each hyperparameter maps directly to a distinct mechanistic function within the search architecture, the consequences of parameter adjustment are comparatively interpretable. In our experiments, the FLEET configuration was derived from a single calibration pass, whereas the temperature-sampling baseline required repeated generations during temperature parameter optimization through Bayesian search.

To systematically identify the optimal intermediate layer for hidden state extraction, we introduce a composite scoring mechanism evaluated over a calibration prompt. For each candidate layer, this heuristic evaluates three primary criteria:

\begin{itemize}
\item Predictive Alignment ($R_{\mathrm{match}}$): The proportion of intermediate token representations -- projected to the vocabulary space via the logit lens technique -- that identically match the model's final output sequence;
\item Clustering Separability ($S_{\mathrm{silhouette}}$): The topological separability of conditional entropy and varentropy into distinct bimodal distributions, quantified by the Silhouette coefficient;
\item Distributional Deviation ($\Delta_{\mathrm{expected}}(L)$): The divergence between the empirical fraction of states exceeding the derived decision threshold and the target theoretical expectation.
\end{itemize}

Synthesizing these metrics, the final layer-selection score $S$ is computed as:

\begin{equation}
S(L) = R_{\mathrm{match}}(L) \cdot S_{\mathrm{silhouette}}(L) - \Delta_{\mathrm{expected}}(L)
\end{equation}

\begin{figure}[t]
\centering
\begin{subfigure}[t]{0.34\linewidth}
  \centering
  \includegraphics[width=\linewidth]{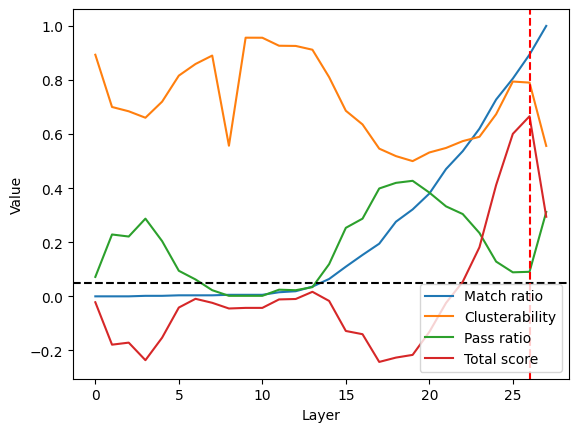}
  \caption{Layer-wise evaluation metrics and composite scoring.}
  \label{fig:image3_1_a}
\end{subfigure}
\hfill
\begin{subfigure}[t]{0.64\linewidth}
  \centering
  \includegraphics[width=\linewidth]{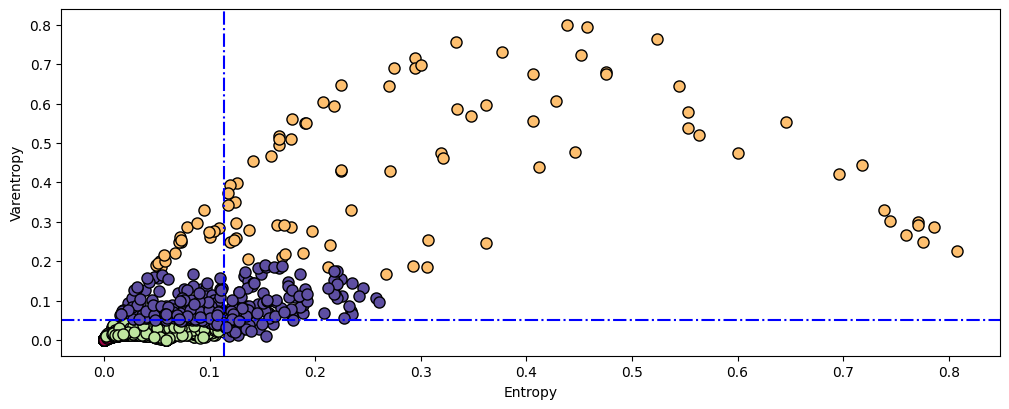}
  \caption{Bivariate entropy--varentropy GMM clustering.}
  \label{fig:image3_1_b}
\end{subfigure}
\par\medskip
\begin{subfigure}[t]{0.44\linewidth}
  \centering
  \includegraphics[width=\linewidth]{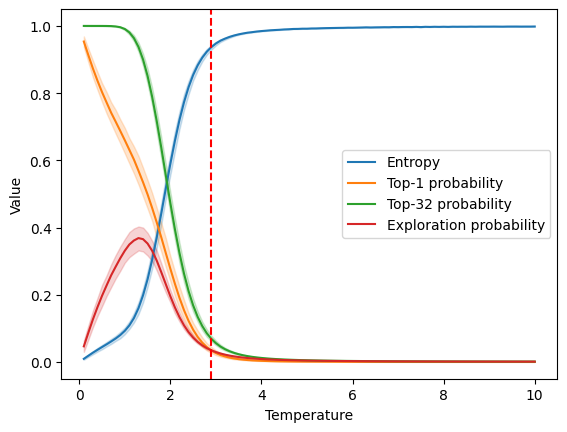}
  \caption{Resampling-temperature calibration.}
  \label{fig:image3_1_c}
\end{subfigure}
\hfill
\begin{subfigure}[t]{0.52\linewidth}
  \centering
  \includegraphics[width=\linewidth]{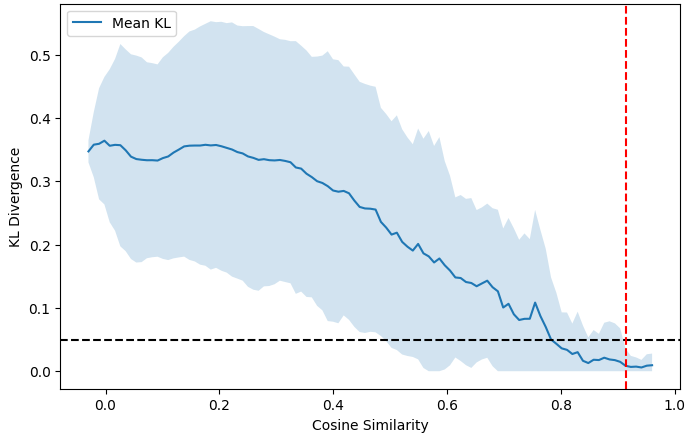}
  \caption{VectorDSU similarity-threshold calibration.}
  \label{fig:image3_1_d}
\end{subfigure}
\caption{Illustrative diagnostics from an example calibration dataset used to refine the FLEET hyperparameter-selection pipeline for Llama 3.2-3B. Task-specific configurations are reported in Table~\ref{tab:hyperparameters}.}
\label{fig:calibration-diagnostics}
\end{figure}

The complete calibration workflow is summarized in Figure~\ref{fig:calibration-diagnostics}.

Figure~\ref{fig:image3_1_a} shows the distinct structural trade-offs observed across the intermediate layers of Llama 3.2-3B. As expected, the logit lens match ratio scales monotonically toward the terminal layers, reflecting progressive convergence onto the output vocabulary. However, an inverse relationship emerges between match ratio and topological clusterability (Silhouette score). Consequently, the final transformer layers prove suboptimal under our selection criteria due to severe degradation in entropy-varentropy separability.

Upon selecting the optimal target layer, we construct an expanded calibration dataset of continuous hidden states and corresponding output logits sampled across multiple prompts.

To parameterize decision boundaries within the joint two-dimensional entropy-varentropy space $(H, V)$, initial thresholds are rederived via bivariate clustering using a Gaussian Mixture Model (GMM) \citep{bishop2006pattern}. GMM density estimation effectively captures complex multi-modal cluster geometries across varying spatial densities. Contrary to our initial hypothesis of a simple quadrant-based partition (delineating high/low regimes of entropy and varentropy), the empirical joint distribution exhibits a cascading cluster topology. Given that layer selection favors independent metric clusterability, extracting decision thresholds from the primary mixture components refines the univariate boundaries by setting them equal to marginal points of corresponding axes on a joint 2D decision frontier.

\FloatBarrier
Figure~\ref{fig:image3_1_b} visualizes the two-dimensional Gaussian Mixture Model clustering applied to the empirical distribution of state entropies and varentropies. The projection captures distinct density gradients that robustly delineate discrete topological regimes. The analytically derived decision thresholds, which partition these states, are demarcated by the blue dash-dotted lines.

Beyond spatial thresholds, the resampling temperature $T_{\mathrm{resample}}$ and the truncation bound $k$ are explicitly parameterized to govern the underlying search dynamics. The candidate window size $k$ is constrained by the expected computational search budget. Subsequently, to impose a structural bias toward early-stage exploration, the temperature scalar $T_{\mathrm{resample}}$ is dynamically calibrated to a critical transition point. Specifically, $T_{\mathrm{resample}}$ is annealed to the minimal value where the cumulative probability mass of the exploratory candidate tail (i.e., the $k-1$ subordinate tokens) strictly outweighs the exploitative probability mass of the single maximum-likelihood candidate (top-1).

Figure~\ref{fig:image3_1_c} illustrates the temperature-dependent evolution of cumulative probability masses across distinct token partitions: the maximum-likelihood candidate (exploitation, top-1), the truncated candidate window (top-$k$, parameterized here as $k=32$), and the exploratory tail (top-$k$ -- top-1). The optimal temperature scalar is empirically identified at the critical intersection point where the exploratory mass strictly surpasses the exploitative mass (observed at $T \approx 2.9$). Furthermore, the plot overlays the corresponding normalized entropy, providing a quantitative measure of how close the annealed distribution approaches a uniform noise prior, thereby guaranteeing that semantic coherence is maintained.

The VectorDSU similarity threshold ($\tau_{\mathrm{dsu}}$) was calibrated by leveraging the spatial cosine similarity-to-divergence correspondence established during the structural formulation. Figure~\ref{fig:image3_1_d} plots the upper confidence bound of this transition divergence (mean plus two standard deviations, $\mu + 2\sigma$) as a function of increasing state proximity. The final operational threshold $\tau_{\mathrm{dsu}}$ (demarcated in red) is formally defined as the minimal similarity value beyond which the normalized divergence is strictly bounded below an error tolerance of 0.05 (demarcated in black).

\FloatBarrier

\section{Experiments}

We empirically validate the proposed FLEET algorithm across two standardized benchmarks characterized by objectively verifiable reward functions: LiveCodeBench \citep{jain2024livecodebench} for algorithmic code generation, and GSM8K \citep{cobbe2021training} for mathematical reasoning. We benchmark the performance of FLEET against standard stochastic temperature sampling across two distinct evaluative paradigms:

\begin{itemize}
\item Ground-Truth Verification: This setting establishes an empirical upper bound by assuming access to deterministic, ground-truth-guided feedback. FLEET is supplied with exact programmatic rewards upon sequence completion: a scalar value of 1.0 for absolute correctness, or a continuous partial reward $r \in [0, 1)$ proportional to the execution correctness or error severity, as evaluated by the benchmark's execution environment. Specifically, the LiveCodeBench reward maps directly to the fraction of passed unit tests, whereas the GSM8K reward constitutes a sparse, binary signal ($r \in \{0, 1\}$).
\item Outcome Reward Model (ORM) Guidance: To simulate a realistic deployment environment without ground-truth access, search trajectories are steered using a surrogate ORM. Under this regime, the ORM acts as the terminal evaluator, supplying FLEET with continuous reward signals normalized to the same $[0,1]$ range as the ground-truth verifier. The underlying scores are derived from parameterized preference distributions learned via pairwise preference optimization.
\end{itemize}

To mitigate data contamination, the LiveCodeBench evaluation is confined to 222 ``easy'' tasks sourced from programming contests held after the cutoff date reported as December 2023; the exact filtering rule and task list are provided in the experiment repository. For the GSM8K evaluation, two demonstrative examples are randomly selected offline from the official training split for each question and then fixed for the corresponding evaluation run.

For every problem, we generate $n=32$ trajectories and report the complete scaling curve for $1\leq k\leq n$. Under ground-truth evaluation, temperature sampling uses the standard combinatorial Pass@$k$ estimator \citep{chen2021evaluating}. If $c$ of the $n$ sampled trajectories are correct, then

\begin{equation}
\widehat{\operatorname{Pass@}k}=1-\frac{\binom{n-c}{k}}{\binom{n}{k}}.
\end{equation}

Because the evaluated FLEET configuration uses deterministic greedy decoding, its prefix-based value at $k$ is 1 if any of the first $k$ trajectories is correct and 0 otherwise. These per-problem values are then averaged across the benchmark.

Consequently, metric calculations are adapted to align with the underlying search mechanisms, particularly when deploying an Outcome Reward Model (ORM) for solution selection:

\begin{itemize}
\item Deterministic Selection (FLEET): Let $y_i\in\{0,1\}$ be the correctness of trajectory $i$ and $r_i$ its ORM score. For the deterministic FLEET prefix, ORM-selected accuracy at $k$ is $y_{j_k}$, where $j_k=\arg\max_{1\leq i\leq k}r_i$.
\item Stochastic Selection (Baseline Temperature Sampling): For the $n$ sampled trajectories, ORM-selected accuracy at $k$ is the average over all size-$k$ subsets $S$:
\begin{equation}
\widehat{A}^{\mathrm{ORM}}_k
=\binom{n}{k}^{-1}
\sum_{\substack{S\subseteq\{1,\ldots,n\}\\|S|=k}}
y_{\arg\max_{i\in S}r_i}.
\end{equation}
This estimates the probability that the ORM's highest-scoring candidate is correct when $k$ candidates are selected from the sampled pool.
\end{itemize}

We employ Llama 3.2-3B as the core base language model because it provides a practical balance among inference throughput, computational requirements, and task performance. The present evaluation is limited to this dense autoregressive architecture; applying FLEET to Mixture-of-Experts (MoE), multimodal, or reflection-heavy reasoning architectures may require architecture-specific calibration.

To ensure structural and distributional representation alignment across the pipeline, we deploy Skywork-Reward-V2 \citep{liu2025skyworkrewardv2} -- which is also fine-tuned on the Llama 3.2-3B architecture -- as our Outcome Reward Model (ORM).

Hyperparameters for FLEET were selected using the automated layer and threshold calibration protocol detailed in Section 3.3. To ensure a rigorous baseline comparison, the sampling temperature $T$ for standard stochastic decoding was independently optimized via Bayesian search over an isolated split of the target benchmarks.

The complete experimental infrastructure was implemented using the Hugging Face Transformers library \citep{wolf2019huggingfaces} combined with nnsight \citep{fiottokaufman2024nnsight} for low-overhead internal-state activation inspection and dynamic logit interventions during the forward pass. Exact checkpoint identifiers, software versions, prompts, and experiment configurations are provided in the linked repository.

All generations are limited to 1,024 new tokens. A sequence that reaches this limit without producing an end-of-sequence token is evaluated exactly as generated and is treated as unfinished; no additional completion or repair step is applied during evaluation. Baseline temperature sampling uses sampling-enabled decoding with the task-specific temperature reported in Table~\ref{tab:hyperparameters}. FLEET uses greedy decoding and applies its penalties to the model's raw, unscaled output logits before token selection. The system-prompt structure was adapted from the prompting setup used by \citet{laban2025llms} and then specialized for the respective coding and mathematical-reasoning tasks. The complete prompts and the GSM8K few-shot construction routine are reproduced in Appendix~\ref{app:prompts}.

\begin{table}[H]
\centering
\begin{tabularx}{\linewidth}{@{}llX@{}}
\toprule
Task & Method & Configuration \\
\midrule
GSM8K & Temperature sampling & $T=0.6$ \\
 & FLEET & $L=26$, $\tau_H=0.11$, $\tau_V=0.05$, $k=32$, $\tau_{\mathrm{dsu}}=0.9$, $T_{\mathrm{resample}}=3.1$ \\
\addlinespace
LiveCodeBench & Temperature sampling & $T=0.8$ \\
 & FLEET & $L=27$, $\tau_H=0.09$, $\tau_V=0.03$, $k=32$, $\tau_{\mathrm{dsu}}=0.9$, $T_{\mathrm{resample}}=1.0$ \\
\bottomrule
\end{tabularx}
\caption{Task-specific hyperparameter configurations for baseline temperature sampling and FLEET. $L$ is the LogitLens layer; $\tau_H$ and $\tau_V$ are entropy and varentropy thresholds; $k$ is the pUCT candidate-window size; $\tau_{\mathrm{dsu}}$ is the VectorDSU similarity threshold; and $T_{\mathrm{resample}}$ is the temperature used to compute the pUCT prior.}
\label{tab:hyperparameters}
\end{table}

\subsection{Empirical Results}

\subsubsection{Ground-Truth Verification}

Table~\ref{tab:ground-truth-results} summarizes the comparative Pass@32 performance of the FLEET algorithm and the baseline temperature sampling method under the ground-truth verification regime. In this setting, the search process receives ground-truth feedback upon the completion of each trajectory.

The empirical results demonstrate that FLEET outperforms standard stochastic sampling across both evaluation domains in this evaluation. The most substantial gains are observed on the LiveCodeBench programming benchmark, where FLEET achieves a 6.31-percentage-point absolute increase in Pass@32 accuracy, resolving 14 additional algorithmic tasks compared to the baseline.

For the GSM8K mathematical reasoning dataset, baseline temperature sampling exhibits performance saturation, successfully resolving 97.27\% of the problem set. Despite this ceiling effect, the FLEET framework maintains a positive advantage in the reported run by successfully decoding 7 additional mathematical tasks that the baseline failed to resolve within the same candidate budget.

\begin{table}[H]
\centering
\begin{tabularx}{\linewidth}{@{}Xrr@{}}
\toprule
Condition & Accuracy (\%) & Problems solved \\
\midrule
\multicolumn{3}{@{}l}{\textbf{GSM8K}} \\
Baseline (temperature sampling) & 97.27 & 1283 \\
FLEET & \textbf{97.80 (+0.53)} & \textbf{1290 (+7)} \\
\addlinespace
\multicolumn{3}{@{}l}{\textbf{LiveCodeBench}} \\
Baseline (temperature sampling) & 59.90 & 133 \\
FLEET & \textbf{66.21 (+6.31)} & \textbf{147 (+14)} \\
\bottomrule
\end{tabularx}
\caption{Pass@32 under ground-truth verification. Parenthesized values are deltas from the baseline within each task; accuracy deltas are percentage-point differences.}
\label{tab:ground-truth-results}
\end{table}

\begin{figure}[t]
\centering
\begin{subfigure}[t]{0.48\linewidth}
  \centering
  \includegraphics[width=\linewidth]{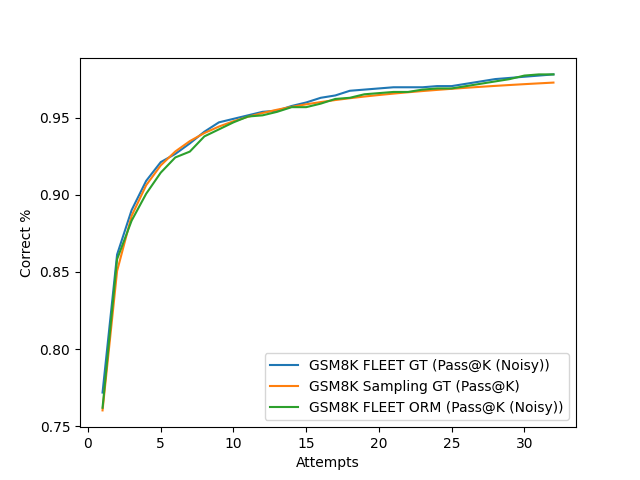}
  \caption{GSM8K.}
  \label{fig:image4_1_a}
\end{subfigure}
\hfill
\begin{subfigure}[t]{0.48\linewidth}
  \centering
  \includegraphics[width=\linewidth]{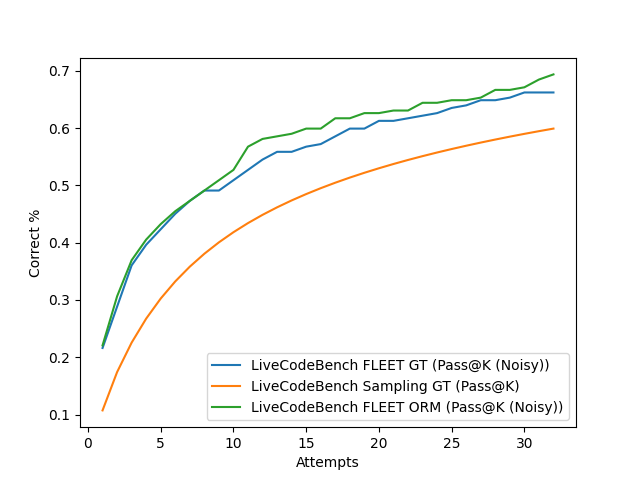}
  \caption{LiveCodeBench.}
  \label{fig:image4_1_b}
\end{subfigure}
\caption{Pass@$k$ accuracy scaling with candidate budget under ground-truth evaluation. Curves labeled ``FLEET ORM'' show ground-truth evaluation of candidates generated with ORM feedback.}
\label{fig:ground-truth-scaling}
\end{figure}
Figure~\ref{fig:image4_1_a} illustrates the scaling efficiency of the decoding strategies as candidate budgets increase on the GSM8K dataset. By systematically enforcing orthogonal search paths via the VectorDSU mechanism, FLEET avoids redundant trajectory exploration and demonstrates better scaling dynamics.

Figure~\ref{fig:image4_1_b} illustrates the comparative scaling behavior of FLEET and baseline temperature sampling across increasing candidate budgets on the LiveCodeBench dataset. Unlike GSM8K where the model shows high base accuracy, programming proves to be more challenging.

In contrast to temperature sampling, FLEET demonstrates a steeper and more robust scaling trajectory.

\subsubsection{Outcome Reward Model (ORM) Guidance}

Table~\ref{tab:orm-results} summarizes the comparative Pass@32 performance under this ORM-guided regime, alongside an auxiliary evaluation -- denoted as FLEET (GT-evaluated) -- that measures the true underlying correctness of the trajectories generated by FLEET when judged by the ground-truth verifier.

LiveCodeBench: Under ORM selection, FLEET outperforms standard temperature sampling, raising the solution rate from 19.36\% (43 problems) to 25.22\% (56 problems). However, comparing the ORM-selected performance (0.2522) against FLEET's ground-truth-evaluated candidate-pool accuracy (0.6937) exposes a substantial reward-ranking gap. This indicates that while the search framework successfully discovers correct algorithmic solutions within its candidate pool, the surrogate reward model struggles to reliably rank them above incorrect alternatives. It also shows that while ORM is not a reliable verifier for LiveCodeBench, its feedback does not make FLEET diverge into reward hacking: the GT-evaluated score does not plateau earlier than the GT-guided score.

GSM8K: On the mathematical reasoning benchmark, baseline temperature sampling marginally outperforms FLEET under ORM selection (0.8544 versus 0.8362). We hypothesize that this occurs because the reward model uses indirect cues for answer ranking -- a byproduct of its conditioning on preference data rather than ground-truth approximation. Crucially, however, the FLEET (GT-evaluated) metric achieves 0.9780 (1290 solved problems), identical to its standalone ground-truth performance.

\begin{table}[H]
\centering
\begin{tabularx}{\linewidth}{@{}Xrr@{}}
\toprule
Condition & Accuracy (\%) & Problems solved \\
\midrule
\multicolumn{3}{@{}l}{\textbf{GSM8K}} \\
Baseline (temperature sampling, ORM-selected) & \textbf{85.44} & \textbf{1127} \\
FLEET, ORM-selected & 83.62 (-1.82) & 1103 (-24) \\
FLEET, ground-truth-evaluated & \emph{97.80 (+12.36)} & \emph{1290 (+163)} \\
\addlinespace
\multicolumn{3}{@{}l}{\textbf{LiveCodeBench}} \\
Baseline (temperature sampling, ORM-selected) & 19.36 & 43 \\
FLEET, ORM-selected & \textbf{25.22 (+5.86)} & \textbf{56 (+13)} \\
FLEET, ground-truth-evaluated & \emph{69.37 (+50.01)} & \emph{154 (+111)} \\
\bottomrule
\end{tabularx}
\caption{Pass@32 using an ORM for final candidate selection, alongside ground-truth evaluation of FLEET-generated trajectories. Parenthesized values are deltas from the baseline within each task.}
\label{tab:orm-results}
\end{table}

\begin{figure}[t]
\centering
\begin{subfigure}[t]{0.48\linewidth}
  \centering
  \includegraphics[width=\linewidth]{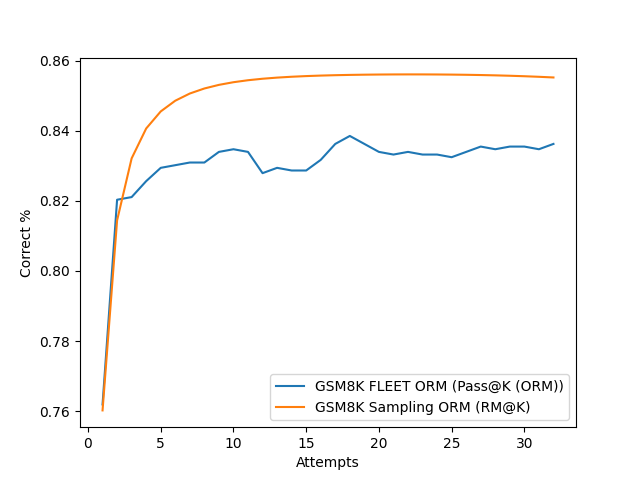}
  \caption{GSM8K.}
  \label{fig:image4_2_a}
\end{subfigure}
\hfill
\begin{subfigure}[t]{0.48\linewidth}
  \centering
  \includegraphics[width=\linewidth]{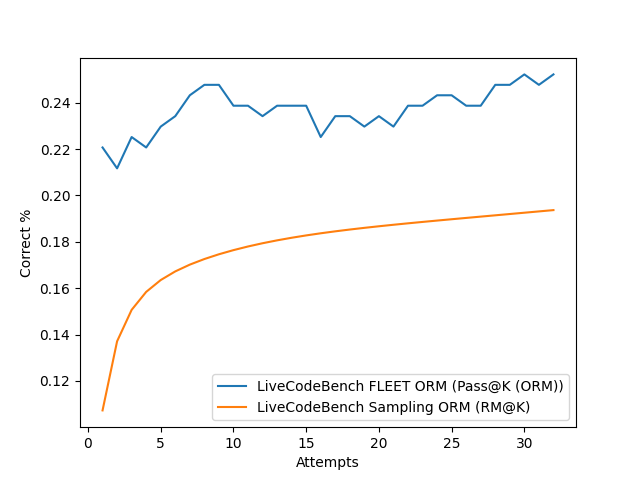}
  \caption{LiveCodeBench.}
  \label{fig:image4_2_b}
\end{subfigure}
\caption{Pass@$k$ accuracy scaling with candidate budget under ORM guidance.}
\label{fig:orm-scaling}
\end{figure}
Figure~\ref{fig:image4_2_a} illustrates the comparative scaling behavior of baseline temperature sampling and FLEET when candidate selection is mediated by an Outcome Reward Model (Skywork-Reward-V2) across increasing candidate budgets on the GSM8K dataset. In contrast to the ground-truth-guided setting, the ORM-evaluated scaling curves reveal a performance plateau and a widening gap relative to true underlying generation capability. This divergence underscores the susceptibility of surrogate reward models to ranking errors when evaluating structurally diverse trajectories. Specifically, because FLEET induces exploration through entropy-varentropy triggers and VectorDSU, the resulting paths may diverge from standard stylistic patterns preferred by the proxy reward model, leading to suboptimal candidate selection at higher candidate budgets.

Figure~\ref{fig:image4_2_b} illustrates the scaling dynamics of FLEET versus baseline temperature sampling on LiveCodeBench when candidate selection is mediated by the Skywork-Reward-V2 outcome reward model. Unlike the performance compression observed in mathematical reasoning, FLEET maintains a sustained and widening performance advantage over stochastic sampling as the candidate budget expands. This suggests that structured, entropy-guided exploration degrades less under imperfect surrogate reward guidance on the more complex task evaluated here.

\subsection{Discussion and Synthesis of Experimental Findings}

A holistic evaluation of the empirical results highlights distinct operational trade-offs and behavioral patterns within the FLEET framework:

\begin{itemize}
\item \textbf{Advantage Under Ground-Truth Verification:} When supplied with deterministic, ground-truth feedback, FLEET outperforms standard stochastic temperature sampling across both evaluated domains. The performance delta is modest on GSM8K due to the already high ceiling, but pronounced on LiveCodeBench, where FLEET resolves 14 additional programming tasks out of 222 total ($\sim6.3$ percentage points). This result is consistent with FLEET suppressing redundant sampling and directing the candidate budget toward distinct reasoning paths.
\item \textbf{The Surrogate Reward Bottleneck:} Coupling search frameworks with an Outcome Reward Model introduces notable performance shifts. Under ORM selection, baseline sampling marginally surpasses FLEET on GSM8K, whereas FLEET retains its superiority on LiveCodeBench, securing a lead of 13 tasks over the baseline despite overall performance degradation across both methods. A key advantage of integrating an ORM within the search loop is continuous reward shaping. While GSM8K's standard verifier provides sparse binary signals, utilizing an ORM populates FLEET's internal memory with smooth, continuous score trajectories. In hybrid configurations -- where search dynamics leverage dense ORM score memory but final evaluation uses ground truth -- FLEET resolves an additional 7 LiveCodeBench tasks within the same candidate budget. This suggests that surrogate models can provide useful dense guidance for internal search routing even when their terminal ranking accuracy is suboptimal.
\end{itemize}

\section{Conclusion}

In this work, we introduced FLEET, a test-time search framework that augments repeated generation with deterministic, state-aware trajectory exploration in its greedy-decoding configuration. By leveraging a dedicated structural memory (VectorDSU) to organize intermediate generation paths and dynamically adjusting exploration via entropy--varentropy trigger regimes, FLEET enhances compute-optimal test-time scaling.

Evaluation across mathematical reasoning and code-generation benchmarks demonstrates the potential of the approach. While maintaining competitive performance under high-baseline saturation on GSM8K, ground-truth-guided FLEET improves LiveCodeBench Pass@32 accuracy from 59.9\% to 66.2\%. When ORM feedback guides the search and the resulting candidate pool is evaluated using ground truth, accuracy reaches 69.4\%, indicating additional candidate-generation potential that the ORM does not reliably recover during final selection. FLEET therefore provides a practical approach for augmenting large language models with structured completion-space search.

\bibliographystyle{plainnat}
\bibliography{library}

\clearpage
\appendix
\section{Prompts and Generation Configuration}
\label{app:prompts}

\subsection{LiveCodeBench system prompt}

The LiveCodeBench problem specification is supplied as the user message following this system prompt:

\begin{lstlisting}
"""
You are an expert Python programmer. You will be given a question (problem specification) and will generate a correct Python program that matches the specification and passes all tests.

Format:
- [Standalone] Make sure that your answer consists of only one Python function at the top level. Do not wrap with a class or split into multiple functions.
"""
\end{lstlisting}

\subsection{GSM8K system prompt}

\begin{lstlisting}
"""
You are a helpful assistant. Your task is to help solving simple math problems. Try to break the problem into substeps, so it is transparent how
you have arrived to the final solution, just like in example QA pairs.

Format:
- Final answer should be a number, not an expression and is always the final line of the solution, preceded by ####.
"""
\end{lstlisting}

\subsection{GSM8K task and few-shot prompt construction}

The following routine constructs the test tasks and independently selects two training examples offline for each question. The constructed prompts are then fixed for the corresponding evaluation run.

\begin{lstlisting}[language=Python]
repo_id = "openai/gsm8k"
gsm_dataset = load_dataset(repo_id, 'main', split='test')
gsm_few_shots = load_dataset(repo_id, 'main', split='train')

math_tasks = [item for item in gsm_dataset]

def clean_answer(question):
    return re.sub(r"<<.*>>", "", question)

def get_few_shot_prompt(examples_count=2):
    random_examples = []

    for _ in range(examples_count):
        example_id = random.randint(1, len(gsm_few_shots)) - 1
        random_examples.append(example_id)

    few_shot_items = gsm_few_shots.select(random_examples)

    few_shot_pieces = []
    for f in few_shot_items:
        few_shot_prompt = f"Question: {f['question']}\nAnswer: {clean_answer(f['answer'])}\n\n"
        few_shot_pieces.append(few_shot_prompt)

    few_shot_prompt = "".join(few_shot_pieces)

    return few_shot_prompt

for i, item in enumerate(math_tasks):
    item['task_id'] = f'gsm8k_{i}'
    item['source'] = 'gsm8k'
    item['prompt'] = get_few_shot_prompt() + f"Question: {item['question']}\nAnswer:"
\end{lstlisting}

\subsection{Decoding configuration}

Both benchmarks use a maximum of 1,024 generated tokens per trajectory. Temperature-sampling baselines enable stochastic sampling and use the temperatures in Table~\ref{tab:hyperparameters}. FLEET disables stochastic sampling, decodes greedily, and receives the generator's raw logits before any temperature scaling. Trajectories that reach the token limit are submitted to the corresponding verifier or ORM in their unfinished form.

\end{document}